\documentclass[twoside]{article}

\usepackage{url}
\usepackage{float}
\usepackage{booktabs}
\usepackage{graphicx}
\usepackage{amssymb}
\usepackage{hyperref}
\usepackage{subcaption}
\usepackage{svg}

\usepackage[preprint]{aistats2027}

\usepackage[round]{natbib}

\begin{document}

\twocolumn[

\aistatstitle{Uncertainty-Aware Optimization for Physics-Aware Highway Trajectory Prediction}

\aistatsauthor{
Aanchal Chugh \And Sebastian Dorn
}

\aistatsaddress{
Technische Hochschule Augsburg \\
TTZ Landsberg am Lech \\
aanchal.rajesh.chugh@tha.de
\And
Technische Hochschule Augsburg \\
TTZ Landsberg am Lech \\
sebastian.dorn@tha.de
}

]
\begin{abstract}
Accurate trajectory forecasting and well-defined predictive uncertainty are crucial for reliable, safety-critical applications such as autonomous driving. Most trajectory prediction approaches provide point estimates only, while uncertainty-aware approaches typically quantify uncertainty only in the trajectory space. In physics-aware approaches, uncertainty in the predicted motion variables should be explicitly modeled and propagated through the vehicle dynamics. Otherwise, the resulting trajectory-space uncertainty may not fully reflect the variability introduced by the underlying motion prediction. Therefore, in this work, uncertainty-aware extensions of X-TRACK (X-TRACK-DE and X-TRACK-MCD), a physics-aware trajectory prediction framework, are proposed. The proposed framework predicts future vehicle motion variables and models both aleatoric and epistemic uncertainties by propagating motion space uncertainty to trajectory space. Additionally, conformal prediction is applied to the trajectory space predictive covariance to construct uncertainty regions targeting a desired marginal coverage level. Evaluation on the highD dataset shows that X-TRACK-DE improves trajectory prediction accuracy over the deterministic baseline, while both uncertainty-aware variants provide predictive uncertainty that can be conformally calibrated to the desired marginal coverage level. 
\end{abstract}

\section{INTRODUCTION}
In the domain of autonomous driving, accurate trajectory prediction of neighboring vehicles is a crucial task to support downstream tasks such as motion planning, collision avoidance, and risk assessment~\citep{bahari_vehicle_2022}. Recent deep learning-based trajectory prediction approaches have shown significant improvements by modeling temporal dependencies, neighboring vehicle interactions, and multi-modal future trajectories. However, in the case of safety-critical applications such as autonomous driving, it is insufficient to have only accurate predictions, as the future motion is uncertain due to interactions among road users, measurement noise, partially observed driver intentions, and scenarios that may be insufficiently represented in the training data~\citep{nayak_pedestrian_2024}. 

Existing neural uncertainty-aware trajectory prediction approaches predominantly estimate uncertainty directly in trajectory space~\citep{nayak_uncertainty_2022, hu_vehicle_2022, liu_uncertainty-aware_2024}. However, when an underlying dynamical system produces the future trajectory, the uncertainty arises not only from the future positions but also from the variables governing the vehicle state's evolution. In this case, the uncertainty over predicted vehicle motion variables could be propagated through such dynamics to capture uncertainty over future trajectories. Previous work proved that uncertainty entering a forecasting pipeline can not be disregarded. It is essential to propagate the uncertain perpetual states through trajectory forecasting. Otherwise, it could lead to overconfident predictions~\citep{ivanovic_propagating_2022}. 

In this paper, an uncertainty-aware extension of X-TRACK~\citep{chugh_x-track_2026} (eXtended LSTM for TRAjectory prediction Constrained by Kinematics) is proposed. \mbox{X-TRACK} is a physics-aware vehicle trajectory framework that decouples the prediction of vehicle motion variables from the future trajectory generation. The model learns to predict the vehicle motion parameters and propagate them through a kinematic layer to obtain vehicle positions. Therefore, \mbox{X-TRACK} is suitable for estimating uncertainties in the vehicular motion space, followed by propagation through the non-linear vehicle dynamics to estimate uncertainty in trajectory space.

In this paper, highway trajectory prediction using the observed motion histories of the target and surrounding vehicles is considered. In this setting, given $3$ s of historical observations, the model predicts $5$ s of future motion without relying on an HD map or additional semantic scene information. In this setting, future motion is inferred from observed vehicle states and interactions without relying on additional map priors, representing a map-independent prediction scenario.

The proposed extension of \mbox{X-TRACK} captures both aleatoric and epistemic uncertainty. The epistemic uncertainty is estimated using Monte Carlo (MC) dropout~\citep{gal_dropout_2016} and deep ensembles~\citep{lakshminarayanan_simple_2017}, whereas the aleatoric uncertainty is represented through heteroscedastic Gaussian distributions over the predicted vehicle motion variables. Samples drawn from these distributions pass through the kinematic layer, allowing uncertainty to evolve over the prediction horizon according to the underlying vehicle dynamics. The main contributions are as follows:
\begin{itemize}
    \item Introduce an uncertainty-aware trajectory prediction framework built on \mbox{X-TRACK} that explicitly models both aleatoric and epistemic uncertainty over predicted vehicle motion variables.
    \item Formulate a probabilistic motion-to-trajectory uncertainty propagation framework to obtain trajectory space uncertainty estimates from motion space.
    \item Evaluate uncertainty modeling, propagation, and conformal prediction with respect to prediction accuracy, calibration, and coverage.
\end{itemize}
\section{RELATED WORK}
This section covers related prior works on physics-aware deep learning networks and the existing uncertainty quantification methods.
\subsection{Physics-aware and Probabilistic Trajectory Prediction}
Integration of a kinematic layer into a deep learning framework enables the model to generate trajectories consistent with the underlying vehicle motion model~\citep{cui_deep_2020}. While many existing trajectory prediction frameworks~\citep{deo_convolutional_2018,messaoud_attention_2021,mo_graph_2021} predict future positions in Cartesian space, physics-aware approaches ~\citep{neumeier_reliable_2024,salzmann_trajectron_2020} incorporate vehicle dynamics to constrain the generated trajectories according to an underlying motion model. This type of hybrid approach usually improves both prediction accuracy and physical feasibility. X-TRACK, a physics-aware highway trajectory prediction framework, explicitly predicts future vehicle motion variables followed by a kinematic rollout to obtain the vehicle position. This separation between the prediction of motion variables and position coordinates is particularly significant for our work as it enables us to propagate uncertainty estimated in motion space to trajectory space.
\subsection{Uncertainty Quantification in Trajectory Prediction}
For uncertainty-aware trajectory prediction, it is essential to produce probabilistic predictions of future motion rather than obtaining a deterministic output~\citep{kahn_uncertainty-aware_2017,wu_motion_2021}. Bayesian Neural Networks have been widely adopted to capture uncertainty in both classification and regression tasks. As exact Bayesian inference is computationally challenging, approximate inference methods such as MC dropout~\citep{gal_dropout_2016} and deep ensembles~\citep{lakshminarayanan_simple_2017} have been developed. Without making substantial changes to the model architecture, these methods have been proven to approximate the posterior and output probabilistic predictions. The uncertainty estimated using these approaches is termed epistemic uncertainty, which arises from the model's lack of knowledge.

The remaining uncertainty comes from the aleatoric component, which is uncertainty inherent in observations and is irreducible. In the case of deep neural networks, heteroscedastic likelihood models are widely used to represent input-dependent aleatoric uncertainty.~\cite{nayak_uncertainty_2022} shows the significance of ensemble disagreement for uncertainty-aware pedestrian trajectory forecasting under sensing uncertainty.~\cite{distelzweig_entropy-based_2024} has focused on uncertainty estimation for trajectory-space prediction (without an explicit physics-based rollout), investigating how uncertainty can be decomposed and evaluated rather than considering the spread of multi-modal predictions as a single notion of confidence.
\subsection{Uncertainty Propagation and Calibration}
As uncertainty is to be captured in the case of a hybrid approach, it is important to understand how uncertainty should propagate through the trajectory prediction pipeline.~\cite{ivanovic_propagating_2022} demonstrated that ignoring uncertain inputs could lead to overconfident predictions, making it essential to propagate upstream perceptual state through trajectory forecasting. Their focus is on the uncertainty associated with the perceived input state. In contrast, our work focuses on uncertainty in predicted future vehicle motion variables and its propagation through the non-linear physics-based kinematic layer, which contributes to uncertainty in trajectory space.

Correspondence between predictive uncertainty and trajectory prediction errors is crucial for reliable uncertainty estimates.~\cite{cao_cctr_2024} proposed a calibration technique specifically for uncertainty-aware motion planning and also proved that the uncertainty estimated by such networks could be poorly calibrated, making a calibration technique significant. Therefore, probabilistic metrics such as negative log-likelihood (NLL), empirical coverage, calibration error, etc. should be considered in addition to trajectory prediction metrics to evaluate the performance of trajectory forecasting models.

In autonomous driving, conformal prediction has recently gained attention for constructing trajectory prediction regions with coverage guarantees for safe planning in dynamic environments~\citep{lindemann_safe_2023}. In this work, underlying aleatoric and epistemic uncertainty is combined with conformal prediction to create prediction regions that target a specified marginal coverage level at each prediction horizon.
\section{BACKGROUND}
This section provides a brief overview of \mbox{X-TRACK}. Instead of explicitly predicting future positions, \mbox{X-TRACK} ~\citep{chugh_x-track_2026}, a physics-based trajectory prediction approach, predicts the vehicle's motion state, followed by a kinematic layer to obtain the vehicle's future trajectory; see Fig.~\ref{fig:uq_track}. Given the observed motion histories of a target vehicle and its surrounding vehicles, \mbox{X-TRACK} independently encodes each vehicle using an xLSTM~\citep{beck_2024_xlstm}. The resulting vehicle-level embeddings are passed to a GAT~\citep{velickovic_graph_2018}, which aggregates the encoded motion information of neighboring vehicles into the target representation using a predefined interaction graph that encodes the spatial vehicle configuration. The LSTM~\citep{hochreiter_1997_lstm} decoder predicts a sequence of future vehicle dynamics: longitudinal acceleration and yaw rate.

Conventionally, trajectory prediction approaches directly predict future positions, whereas \mbox{X-TRACK} generates future trajectories by propagating the predicted vehicle motion variables through a kinematic bicycle model~\citep{polack_kinematic_2017}. The input to the \mbox{X-TRACK} model is the history of a target vehicle along with its neighboring vehicles observed for $t_\mathrm{obs}$ and is defined as
\begin{equation}
\mathbf{X}^{(i)}
=
\left[
\mathbf{u}_{1}^{(i)},
\mathbf{u}_{2}^{(i)},
\ldots,
\mathbf{u}_{t_{\mathrm{obs}}}^{(i)}
\right],
\end{equation}
where $\mathbf{u}_t^{(i)} = [\, a_{x, t}^{(i)}, \dot{\psi}_t^{(i)}\,]^\top$ represents the motion variable input of vehicle $i$ at time step $t$. Here, $a_x$ and $\dot{\psi}$ are the longitudinal acceleration and yaw rate, respectively. The decoder predicts the motion state vector for $t_f$ future time steps
\begin{equation}
\hat{\mathbf{U}}
=
\left[
\hat{\mathbf{u}}_{t_{\mathrm{obs}}+1},
\hat{\mathbf{u}}_{t_{\mathrm{obs}}+2},
\ldots,
\hat{\mathbf{u}}_{t_{\mathrm{obs}}+t_f}
\right],
\end{equation}
where $\hat{\mathbf{u}}_t = [\, \hat{a}_{x,t},\, \hat{\dot{\psi}}_t\,]^\top$. Let the current state of the vehicle at time $t$ be denoted by $\mathbf{s}_t = [x_t,\, y_t,\, v_t,\, \psi_t]^\top,$
where $(x_t,y_t)$ represents the vehicle's position, $v_t$ is the velocity, and $\psi_t$ is the heading angle. \\
The future trajectory is obtained by recursively updating the vehicle state using the underlying physical laws of motion $\mathbf{s}_{t}=f(\mathbf{s}_{t-1}, \mathbf{u}_t),$ where $f(\cdot)$ is the kinematic model used for trajectory rollout. This yields $\mathbf{p}_t = [\, x_{t},\, y_t\,]^\top,$ giving the vehicle's position coordinates at time $t$. Therefore, the future trajectory would be:
\begin{equation}
\mathbf{P}
=
\left[
\mathbf{p}_{t_{\mathrm{obs}}+1},
\mathbf{p}_{t_{\mathrm{obs}}+2},
\ldots,
\mathbf{p}_{t_{\mathrm{obs}}+t_f}
\right].
\end{equation}
\mbox{X-TRACK} explicitly separates the prediction of vehicle motion variables and the physics-based trajectory rollout. This separation enables propagation of prediction uncertainty from vehicle motion variables to positions, forming the basis of the uncertainty propagation framework proposed in this paper. On the highD benchmark, X-TRACK has demonstrated competitive performance relative to the highway trajectory prediction baselines evaluated in ~\cite{chugh_x-track_2026}.

\section{METHODOLOGY}
\label{sec:methodology}
In this section, the modeling of aleatoric and epistemic uncertainties is described, followed by propagation of vehicle motion uncertainties via physics-based rollout and calibration using conformal prediction.
\subsection{Predictive Uncertainty Modeling}
Vehicle trajectory prediction is a supervised regression problem where a neural network predicts a deterministic future trajectory. As \mbox{X-TRACK} is trained with deterministic regression objectives, only point estimates of future vehicle motion variables are generated without a measure of prediction confidence. In safety-critical scenarios, however, the model must also quantify the associated predictive uncertainty to support reliable, risk-aware decision-making~\citep {kahn_uncertainty-aware_2017}.

Aleatoric uncertainty (also called statistical uncertainty) reflects the intrinsic stochasticity of future vehicle behavior that is inherent in the data-generating process~\citep{davis_quantifying_2020}. Epistemic uncertainty (also known as model uncertainty) captures uncertainty in the model and its predictions due to lack of knowledge~\citep{davis_quantifying_2020}. The total predictive uncertainty can be decomposed into aleatoric and epistemic components through the law of total covariance~\citep{nayak_pedestrian_2024}:
$\boldsymbol{\Sigma}_{\mathrm{total}, t}
=
\boldsymbol{\Sigma}_{\mathrm{ale}, t}
+
\boldsymbol{\Sigma}_{\mathrm{epi}, t}.$

The following subsections describe the approaches for estimating epistemic uncertainty, followed by the proposed modeling of aleatoric uncertainty in the predicted vehicle motion variables. 
\subsubsection{Epistemic Uncertainty}
In Bayesian inference, epistemic uncertainty is captured by a posterior distribution over the model parameters. As accurate Bayesian inference is often computationally intractable for deep neural networks, practical approximation methods such as MC Dropout and Deep Ensembles have become widely known for estimating prediction uncertainty~\citep{gawlikowski_survey_2022}.

\textbf{MC Dropout.}~\cite{gal_dropout_2016} introduced dropout as a posterior approximation by performing multiple stochastic forward passes, keeping dropout active during inference. Given the input $\mathbf{X}$, the $M$ stochastic forward passes during inference generate $M$ predicted trajectories, given by
\begin{equation}
\mathcal{P}_{\mathrm{mc}}
=
\left\{
\mathbf{P}^{(m)}
\right\}_{m=1}^{M},
\end{equation}
where $\mathbf{P}^{(m)}$ is the $m^\mathrm{th}$ future trajectory obtained from $\hat{\mathbf{U}}^{(m)}$ after the kinematic rollout, where $\mathbf{p}_{t}^{(m)} = [x_{t}^{(m)},\, y_{t}^{(m)}]^\top$.
The predictive mean $\boldsymbol{\mu}_{\mathrm{mc},t}$ is estimated as the arithmetic mean, while the epistemic covariance is computed as
\begin{equation}
\boldsymbol{\Sigma}_{\mathrm{mc},t}
=
\frac{1}{M}
\sum_{m=1}^{M}
\left(
\mathbf{p}_{t}^{(m)}
-
\boldsymbol{\mu}_{\mathrm{mc},t}
\right)
\left(
\mathbf{p}_{t}^{(m)}
-
\boldsymbol{\mu}_{\mathrm{mc},t}
\right)^{\top}.
\label{eq:mc_covariance}
\end{equation}
The resulting $2\times2$ positional covariance reflects the model's sensitivity to uncertainty in the learned parameters.

\textbf{Deep Ensembles.}~\cite{lakshminarayanan_simple_2017} estimate the epistemic uncertainty by training $K$ independently initialized neural networks on the same data. Every model generates an independent prediction $\mathbf{P}^{(k)}$:
\begin{equation}
\mathcal{P}_{\mathrm{de}}
=
\left\{
\mathbf{P}^{(k)}
\right\}_{k=1}^{K},
\end{equation}
from which the predictive mean and covariance for deep ensembles are computed analogously to MC Dropout. Epistemic uncertainty in deep ensembles arises from disagreement between the ensemble members, as each network converges to a different local optimum. 
\subsection{Aleatoric Uncertainty}
Driver intentions, sensor noise, interactions with neighboring vehicles, etc., impact future motion variables of the vehicle. These irreducible sources of uncertainty contribute to aleatoric uncertainty. 

The proposed framework predicts a probability distribution over future vehicle motion variables to estimate the aleatoric uncertainty~\citep{hullermeier_aleatoric_2021}. The prediction head of \mbox{X-TRACK} is modified to output both the mean and variance of the future vehicle motion variables. Therefore, instead of predicting $\hat{\mathbf{u}}_t$, the model predicts $\hat{\boldsymbol{\theta}}_t$, where 
\begin{equation}
\hat{\boldsymbol{\theta}}_t
=
\left[
\mu_{a_x,t},
\mu_{\dot{\psi},t},
\sigma^2_{a_x,t},
\sigma^2_{\dot{\psi},t}
\right]^\top .
\end{equation}

\subsection{Physics-based Uncertainty Propagation}
As the mapping from vehicle motion variables to vehicle position coordinates is non-linear, the uncertainty estimated in motion space needs to be propagated through the kinematic rollout.

For each time step $t$, the longitudinal acceleration $a_{x,t}$ and yaw rate $\dot{\psi}_t$ are drawn from the predicted Gaussian distribution $a_{x,t} \sim \mathcal{N}\left(\mu_{a_{x,t}},\,\sigma_{a_{x,t}}^{2}\right),$ and
$\dot{\psi}_t \sim \mathcal{N}\left(\mu_{\dot{\psi}_t},\,\sigma_{\dot{\psi}_t}^{2}\right).$ Here, longitudinal acceleration and yaw rate are modeled using diagonal heteroscedastic Gaussian distributions. Therefore, cross-covariance between the predicted motion variables is not explicitly modeled. However, as the nonlinear kinematic rollout could induce correlation between the $x, y$ positions, the final trajectory space uncertainty is represented using a full $2\times2$
positional covariance matrix. The samples are drawn independently to have $N$ different longitudinal accelerations and yaw rates for each time step $t$: $\left\{
a_{x,t}^{(n)},\,
\dot{\psi}_t^{(n)}
\right\}_{n=1}^{N}.$

The state $\mathbf{s}_{t-1}^{(n)}$ for the $n^{th}$ sample drawn from the predicted distribution is passed through the kinematic model to compute the next state $\mathbf{s}_{t}^{(n)} = f(\mathbf{s}_{t-1}^{(n)}, \mathbf{u}_t^{(n)}),$ where $f(\cdot)$ is the kinematic layer of the \mbox{X-TRACK} model~\citep{chugh_x-track_2026}. Hence, each sample drawn from the predicted distribution generates one complete trajectory, given by: $
\mathcal{P}_{\mathrm{ale}}
=
\left\{
\mathbf{P}^{*(n)}
\right\}_{n=1}^{N},$
where $n^{\mathrm{th}}$ trajectory is
\begin{equation}
\mathbf{P}^{*(n)}
=
\left[
\mathbf{p}_{t_{\mathrm{obs}}+1}^{*(n)},
\mathbf{p}_{t_{\mathrm{obs}}+2}^{*(n)},
\ldots,
\mathbf{p}_{t_{\mathrm{obs}}+t_f}^{*(n)}
\right].
\end{equation}
With $\mathbf{p}_{t}^{*(n)} = [x_{t}^{*(n)},\, y_{t}^{*(n)}]^\top$, the mean trajectory is the arithmetic mean of the trajectories generated through aleatoric sampling, and the corresponding positional covariance is estimated from their deviations around this mean. The predicted mean motion variables are propagated through the kinematic layer to obtain a deterministic mean-control trajectory:
\begin{equation}
\mathbf{P}_{\mathrm{det}}
=
f_{\mathrm{rollout}}
\left(
\mathbf{s}_{t_{\mathrm{obs}}},
\left\{
\boldsymbol{\mu}_{\mathbf{u},t}
\right\}_{t=t_{\mathrm{obs}}+1}^{t_{\mathrm{obs}}+t_f}
\right),
\label{eq:det_traj}
\end{equation}
where $\boldsymbol{\mu}_{\mathbf{u},t} = [\mu_{a_{x,t}},\, \mu_{\dot{\psi}_t}]^\top.$ Here, $f_\mathrm{rollout}(\cdot)$ is the kinematic rollout which is applied iteratively from time step $t_\mathrm{obs}+1$ to $t_\mathrm{obs}+t_f$.
\subsection{Combined Uncertainty} 
To estimate combined uncertainty, i.e., both epistemic and aleatoric, a nested approach is followed. In the case of MC dropout, $M$ stochastic forward passes are performed with dropout enabled during inference. For the $m^\mathrm{th}$ stochastic forward pass, the model predicts the vehicular motion distributions at every future time step:
\begin{equation}
   \boldsymbol{\theta}^{(m)}_t
=
\left[
\mu^{(m)}_{a_x,t},
\mu^{(m)}_{\dot{\psi},t},
\sigma^{2(m)}_{a_x,t},
\sigma^{2(m)}_{\dot{\psi},t}
\right]^\top . 
\end{equation}
For each stochastic pass $m$, $N$ different vehicle motion variables are sampled according to
\begin{equation}
a_{x,t}^{(m,n)} \sim \mathcal{N}\left(\mu^{(m)}_{a_{x,t}},\,\sigma_{a_{x,t}}^{2(m)}\right), 
\end{equation}
\begin{equation}
\dot{\psi}_t^{(m,n)} \sim \mathcal{N}\left(\mu^{(m)}_{\dot{\psi}_t},\,\sigma_{\dot{\psi}_t}^{2(m)}\right),
\end{equation}
where $m=1,2,\ldots,M$ and $n=1,2,\ldots,N$. Each sampled vehicle motion state is propagated through the kinematic layer to obtain the vehicle's future coordinates $
\mathbf{s}_{t}^{(m,n)} = f(\mathbf{s}_{t-1}^{(m,n)}, \mathbf{u}_t^{(m, n)}),
$ where $\mathbf{u}_{t}^{(m,n)}
=
[a_{x,t}^{(m,n)},\,
\dot{\psi}_{t}^{(m,n)}]^{\top}$
resulting in $M \times N$ trajectories. At time step $t$, the vehicle position is represented by 
$\mathbf{p}_t^{(m, n)} = [x_{t}^{(m,n)} , y_{t}^{(m,n)}]^\top.$

For each trajectory generated from the $M$ stochastic forward passes along with $N$ aleatoric samples drawn from the predicted motion distribution, the conditional mean position $\bar{\mathbf{p}}_{t}^{(m)}$ is computed as the arithmetic mean, and the predictive mean trajectory is
\begin{equation}
\bar{\mathbf{p}}_{t}^{*}
=
\frac{1}{M}
\sum_{m=1}^{M}
\bar{\mathbf{p}}_{t}^{(m)}
=
\frac{1}{MN}
\sum_{m=1}^{M}
\sum_{n=1}^{N}
\mathbf{p}_{t}^{(m,n)}.
\label{eq:mean_traj}
\end{equation}
The aleatoric covariance is computed by averaging the within-model covariance over all stochastic forward passes as well as sampled vehicle motion variables:
\begin{equation}
\begin{aligned}
\boldsymbol{\Sigma}_{\mathrm{ale},t}
&=
\frac{1}{M}
\sum_{m=1}^{M}
\Bigg[
\frac{1}{N}
\sum_{n=1}^{N}
\left(
\mathbf{p}^{(m,n)}_t-\bar{\mathbf{p}}^{(m)}_t
\right)
\\
&\qquad\qquad
\left(
\mathbf{p}^{(m,n)}_t-\bar{\mathbf{p}}^{(m)}_t
\right)^{\top}
\Bigg].
\end{aligned}
\label{eq:aleatoric_cov}
\end{equation}
Conditioning on a fixed model realization, $\boldsymbol{\Sigma}_{\mathrm{ale},t}$ captures the expected covariance resulting from the stochasticity of the predicted vehicle motion variables. The disagreement between the conditional mean trajectories generated by the $M$ stochastic forward passes provides the estimation of the epistemic covariance 
\begin{equation}
\boldsymbol{\Sigma}_{\mathrm{epi},t}
=
\frac{1}{M}
\sum_{m=1}^{M}
\left(
\bar{\mathbf{p}}^{(m)}_t-\bar{\mathbf{p}}^{*}_t
\right)
\left(
\bar{\mathbf{p}}^{(m)}_t-\bar{\mathbf{p}}^{*}_t
\right)^{\top}.
\label{eq:epistemic_cov}
\end{equation}
Therefore, the total predictive covariance $\boldsymbol{\Sigma}_{\mathrm{total},t}$ in trajectory space, followed by the law of total covariance, is given by $
\boldsymbol{\Sigma}_{\mathrm{total},t}
=
\boldsymbol{\Sigma}_{\mathrm{ale},t}
+
\boldsymbol{\Sigma}_{\mathrm{epi},t},
\label{eq:total_cov}
$
where
\begin{equation}
\boldsymbol{\Sigma}_{\mathrm{total},t}
=
\begin{bmatrix}
\sigma_{x,t}^{2} & \sigma_{xy,t} \\
\sigma_{xy,t} & \sigma_{y,t}^{2}
\end{bmatrix}.
\label{eq:full_covariance}
\end{equation}
The covariance matrix captures correlations between longitudinal and lateral positional uncertainty that may arise through the nonlinear kinematic rollout.
\subsection{Conformal Prediction}
The estimation of aleatoric and epistemic uncertainty does not necessarily guarantee that the resulting trajectory space uncertainty regions are calibrated to a desired coverage level. Therefore, conformal prediction (CP)~\citep{angelopoulos_gentle_2022} is employed to calibrate the uncertainty regions obtained from the underlying uncertainties. 

Let $S_{\mathrm{cal}}$ be the number of calibration scenarios. For
calibration scenario $i$ and prediction time step $t$, the positional residual
is $\mathbf{e}_{t,i}
=
\mathbf{p}_{t,i}
-
\boldsymbol{\mu}_{t,i},$
where $\mathbf{p}_{t,i}\in\mathbb{R}^{2}$ and
$\boldsymbol{\mu}_{t,i}\in\mathbb{R}^{2}$ are the ground-truth position and the corresponding predictive
mean, respectively. 

The conformal score is defined using the Mahalanobis distance under the
total positional covariance: $
r_{t,i}
=
\sqrt{
\mathbf{e}_{t,i}^{\top}
\left(
\boldsymbol{\Sigma}_{\mathrm{total},t,i}
+
\epsilon\mathbf{I}
\right)^{-1}
\mathbf{e}_{t,i}
},
\qquad
\epsilon=10^{-8}.
\label{eq:conformal_score}
$ For every prediction horizon $t$, the calibration scores are collected as
$
\mathcal{R}_{t}
=
\left\{
r_{t,i}
\right\}_{i=1}^{S_{\mathrm{cal}}}.$

For a desired coverage level $1-\delta$, the finite-sample conformal rank is
$
j
=
\left\lceil
\left(S_{\mathrm{cal}}+1\right)(1-\delta)
\right\rceil ,
\label{eq:conformal_rank}
$
and the corresponding horizon-wise conformal quantile is $
q_t=r_{t,(j)},$
where $r_{t,(j)}$ denotes the $j$th smallest score in $\mathcal{R}_t$.

At test time, the conformalized total covariance is defined as
\begin{equation}
\hat{\boldsymbol{\Sigma}}_{\mathrm{cp},t,s}
=
q_t^2
\left(
\boldsymbol{\Sigma}_{\mathrm{total},t,s}
+
\epsilon\mathbf{I}
\right).
\label{eq:conformal_covariance}
\end{equation}

The corresponding conformal prediction region is
\begin{equation}
\mathcal{C}_{t,s}
=
\left\{
\mathbf{p}:
\left(
\mathbf{p}-\boldsymbol{\mu}_{t,s}
\right)^{\top}
\hat{\boldsymbol{\Sigma}}_{\mathrm{cp},t,s}^{-1}
\left(
\mathbf{p}-\boldsymbol{\mu}_{t,s}
\right)
\leq 1
\right\}.
\label{eq:conformal_region}
\end{equation}

A separate conformal quantile is estimated for each prediction horizon,
allowing the prediction regions to account for the increase in prediction
error and uncertainty over time. 

\begin{figure*}[t]
    \centering
    \includegraphics[width=\textwidth]{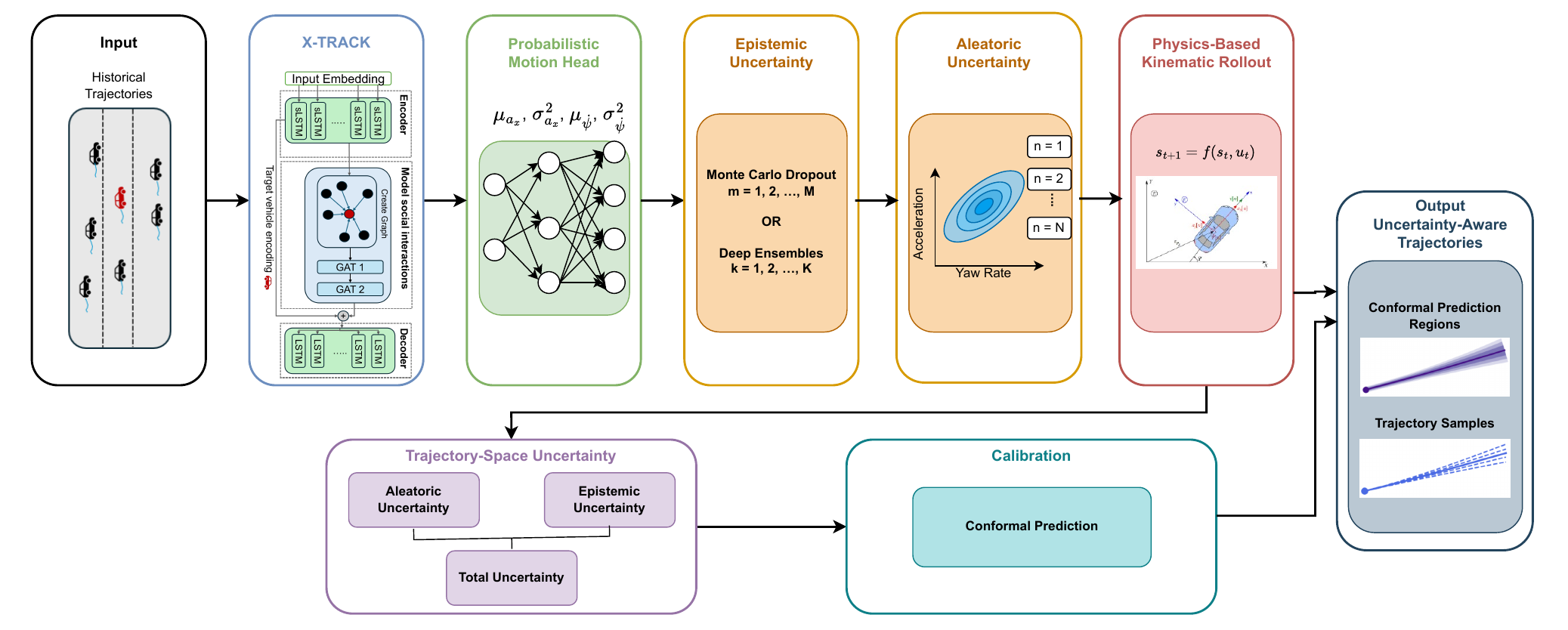}
    \caption {\textbf{Overview of the uncertainty-aware \mbox{X-TRACK} framework.} Aleatoric uncertainty is estimated by sampling from the predicted motion distributions, and epistemic uncertainty is obtained using MC Dropout or Deep Ensembles. The sampled vehicle motion variables are propagated through the physics-based kinematic layer to obtain uncertainty in trajectory space. To obtain calibrated trajectory uncertainty regions, conformal prediction is applied.}
    \label{fig:uq_track}
\end{figure*}

Figure \ref{fig:uq_track} shows the uncertainty-aware extension of \mbox{X-TRACK} ~\citep{chugh_x-track_2026}, where a probabilistic vehicle motion head is employed. The motion head predicts the mean and variance of the longitudinal acceleration and yaw rate, representing a distribution over the vehicle motion controls. From the predicted motion distribution, aleatoric sampling is performed along with MC Dropout or Deep Ensembles to capture both aleatoric and epistemic uncertainty. The samples drawn are propagated through the physics-based kinematic layer to obtain the uncertainties in trajectory space. As a post-hoc step, conformal prediction is applied to construct uncertainty regions targeting the desired marginal coverage level.
\section{EXPERIMENTS}
\label{sec:experiments}
This section provides an overview of the dataset, evaluation metrics, and implementation details. 
\subsection{Dataset}
The experiments are performed on the highD~\citep{krajewski_highd_2018} dataset, which has a sample rate of $f = 25$ Hz, captured using a drone at six different highway locations in Germany. The data preprocessing is performed similarly to \mbox{X-TRACK} to ensure a fair comparison to the uncertainty-aware extension framework proposed in this paper. 

In order to perform conformal prediction, the validation set is split in half to create a calibration dataset, keeping equal proportions of scenarios. The training, validation, calibration, and test ratios are $70:5:5:20$, with $9604$, $686$, $685$, and $2747$ samples, respectively. Each traffic scenario spans $8$s, with $3$s of observed history and $5$s of future trajectory. 
\subsection{Evaluation Metrics and Implementation Details}
\label{sec:eval_metrics}
The definitions of the widely adopted metrics for evaluation of trajectory prediction and uncertainty quantification are described in Appendix \ref{sec:traj_pred_metrics} and \ref{sec:unc_eval_metrics}. The detailed information on model training, hyperparameters, loss function, and settings for the uncertainty quantification, evaluated models, along with information on licenses of the existing assets, can be found in Appendix \ref{sec:baseline} and \ref{sec:train_dets}. Preliminary experiments with additional horizon-wise variance scaling showed negligible changes after conformal calibration because multiplicative scaling is largely absorbed by the normalized conformal score and corresponding quantile. Therefore, this additional scaling step is not used in the final framework.
\section{RESULTS AND DISCUSSION}
\begin{table}[t]
\centering
\caption{ADE and FDE (in meters) evaluated at a $5s$ prediction horizon for the evaluated models}
\label{tab:trajectory_performance_table}
\small
\setlength{\tabcolsep}{2pt}
\resizebox{\columnwidth}{!}{
\begin{tabular}{lcccc}
\toprule
\textbf{Architecture} &
\textbf{ADE} $\downarrow$ &
\textbf{FDE} $\downarrow$ &
\textbf{minADE} $\downarrow$ &
\textbf{minFDE} $\downarrow$ \\
\midrule

X-TRACK~\citep{chugh_x-track_2026}
& 0.56 & 1.76 & - & - \\

\mbox{X-TRACK-MCD} (Ours)
& 0.54 & 1.79 & 0.20 & 0.41 \\

\mbox{X-TRACK-DE} (Ours)
& \textbf{0.44} & \textbf{1.49} & \textbf{0.15} & \textbf{0.33} \\

\midrule

MHA-LSTM~\citep{messaoud_attention_2021}
& 1.97 & 4.72 & - & - \\

iNATran~\citep{chen_vehicle_2022}
& 1.84 & 3.95 & - & - \\

GFTNNv2~\citep{neumeier_prediction_2023}
& 0.92 & 2.20 & - & - \\

cVMD \citep{neumeier_reliable_2024} 
& 1.74 & 4.95 & - & - \\

cVMDx~\citep{neumeier_uq_2026}
& 1.34 & 3.80 & 0.66 & 1.70 \\

X-TRAJ~\citep{chugh_x-track_2026}
& 1.14 & 2.65 & - & - \\

\bottomrule
\end{tabular}
}
\vspace{1mm}
\begin{minipage}{0.98\linewidth}
\footnotesize
\textit{Note:} minADE and minFDE are reported only for methods for which multiple sampled trajectories were evaluated in this study. Therefore, these metrics are unavailable for the remaining baselines.
\end{minipage}
\end{table}
In this section, \mbox{X-TRACK} and its uncertainty-aware versions (Deep Ensembles and MC Dropout) are compared in terms of evaluation metrics. Both uncertainty-aware versions capture aleatoric and epistemic uncertainty along with calibration using conformal prediction. The major difference in the two versions lies in how the epistemic uncertainty is estimated. Therefore, \mbox{X-TRACK} with MC dropout as an epistemic uncertainty approach is termed \mbox{X-TRACK-MCD}, and with deep ensembles is termed \mbox{X-TRACK-DE}.

\textbf{Trajectory Prediction Performance.} Table \ref{tab:trajectory_performance_table} compares the deterministic \mbox{X-TRACK} baseline and its uncertainty-aware variants in terms of the trajectory prediction metrics. \mbox{X-TRACK-DE} achieves the best trajectory accuracy, reducing ADE from $0.56$ to $0.44$ and FDE from $1.76$ to $1.49$. Over the baseline, \mbox{X-TRACK-DE} has approximately $21.4\%$ and $15.3\%$ relative improvements in ADE and FDE, respectively. \mbox{X-TRACK-MCD} achieves a minor improvement in ADE from $0.56$ to $0.54$ but achieves a slightly higher FDE than the baseline. This indicates that toward the end of the prediction horizon, its mean prediction becomes slightly less accurate.

While comparing the minADE@$80$ and minFDE@$80$, \mbox{X-TRACK-DE} and \mbox{X-TRACK-MCD} yield a minADE of $0.15$ and $0.20$, indicating that the trajectories generated from both methods are substantially closer to the ground truth trajectory than their respective predictive means. The lowest minFDE has been achieved by \mbox{X-TRACK-DE} ($0.33$), as compared to \mbox{X-TRACK-MCD} ($0.41$). This demonstrates that there is at least one trajectory generated by deep ensembles with a more accurate position at the final time step.

\begin{table}[t]
\centering
\caption{Comparison of the proposed variants with the baseline models evaluated in the considered highway trajectory prediction setting in terms of RMSE at different prediction horizons.}
\label{tab:rmse_horizon_table}
\small
\setlength{\tabcolsep}{3pt}

\resizebox{\columnwidth}{!}{
\begin{tabular}{lccccc}
\toprule
\textbf{Architecture} &
\textbf{1s} &
\textbf{2s} &
\textbf{3s} &
\textbf{4s} &
\textbf{5s} \\
\midrule

X-TRACK~\citep{chugh_x-track_2026}
& 0.10 & 0.31 & 0.71 & 1.31 & 2.16 \\

\mbox{X-TRACK-MCD} (Ours)
& 0.06 & 0.22 & 0.66 & 1.34 & 2.23 \\

\mbox{X-TRACK-DE} (Ours)
& \textbf{0.04}
& \textbf{0.16}
& \textbf{0.51}
& \textbf{1.09}
& \textbf{1.87} \\

\midrule

MHA-LSTM~\citep{messaoud_attention_2021}
& 0.71 & 1.62 & 2.85 & 4.31 & 6.06 \\

iNATran~\citep{chen_vehicle_2022}
& 0.88 & 1.62 & 2.35 & 3.58 & 4.95 \\

GFTNNv2~\citep{neumeier_prediction_2023}
& 0.47 & 0.61 & 1.05
& 1.75 & 2.69 \\

cVMD~\citep{neumeier_reliable_2024} 
& 0.26  & 1.03  & 2.27  
& 3.93  & 5.99 \\

cVMDx~\citep{neumeier_uq_2026}
& 0.20 & 0.78 & 1.71 & 2.93 & 4.43 \\

X-TRAJ~\citep{chugh_x-track_2026} & 0.48 & 1.01 & 1.58 & 2.20 & 3.17 \\
\bottomrule
\end{tabular}
}
\end{table}
\textbf{Horizon-wise Prediction Performance.} Table \ref{tab:rmse_horizon_table} reports the RMSE at each prediction horizon for \mbox{X-TRACK}  along with its uncertainty-aware variants. The RMSE increases with the prediction horizon for all the methods. \mbox{X-TRACK-DE} achieves the lowest RMSE across all horizons among the evaluated methods, reducing the baseline from $0.10$ to $0.04$ at $1s$ and from $2.16$ to $1.87$ at $5s$. This corresponds to a relative improvement of $13.4\%$ over the \mbox{X-TRACK}  baseline at the $5s$ horizon. Compared to the baseline, \mbox{X-TRACK-MCD} reduces error at shorter horizons, up to $3s$, and yields a higher error than the baseline at $4s$ and $5s$, with RMSE values of $1.34\,\mathrm{m}$ and $2.23\,\mathrm{m}$.
\begin{table}[t]
\centering
\caption{Uncertainty quantification and calibration performance of the evaluated uncertainty-aware trajectory prediction approaches.}
\label{tab:uq_results}
\small
\renewcommand{\arraystretch}{1.08}
\setlength{\tabcolsep}{5pt}
\resizebox{\columnwidth}{!}{
\begin{tabular}{@{}lccc@{}}
\toprule
\textbf{Metric} &
\shortstack{\textbf{X-TRACK}\\\textbf{MCD}} &
\shortstack{\textbf{X-TRACK}\\\textbf{DE}} &
\textbf{cVMDx} \\
\midrule

NLL $\downarrow$
& \textbf{0.47}
& 0.51
& 11.32 \\

ECE $\downarrow$
& 0.17
& 0.12
& \textbf{0.10} \\

Pearson Correlation ($\rho$) $\uparrow$
& \textbf{0.34}
& 0.33
& 0.05 \\

Mean Predictive Uncertainty ($\mathrm{m}^2$)
& 1.32
& 0.75
& 4.04 \\

Miss Rate (MR) @ $2\,\mathrm{m}$ $\downarrow$
& 0.02
& \textbf{0.01}
& 0.31 \\

\midrule

CP Coverage@$95\%$
& 95.72\%
& 95.39\%
& - \\

MPIW@95\% (after CP) $\downarrow$
& 2.61
& \textbf{2.11}
& - \\

ECE (after CP) $\downarrow$
& \textbf{0.11}
& \textbf{0.11}
& - \\

\bottomrule
\end{tabular}
}
\end{table}

\textbf{Uncertainty Quantification Performance.} Table~\ref{tab:uq_results} reports predictive-distribution and
conformal-set metrics (see Appendix~\ref{sec:unc_eval_metrics}) for the evaluated uncertainty-aware approaches. NLL, ECE, Pearson correlation, mean predictive uncertainty, and MR are evaluated prior to conformal calibration, whereas CP coverage and MPIW characterize the conformal prediction regions. Both approaches achieve approximately the same coverage ($95\%$). X-TRACK-DE achieves a lower miss rate ($0.01$ vs.\ $0.02$) compared to \mbox{X-TRACK-MCD}, indicating more frequent final-position hits within $2\,\mathrm{m}$. In addition, the Pearson correlation between ADE and predictive uncertainty is similar: $0.34$ for \mbox{X-TRACK-MCD} and $0.33$ for X-TRACK-DE, indicating a moderate positive correlation between error and uncertainty. The lowest ECE is achieved by cVMDx followed by X-TRACK-DE.

In terms of estimated uncertainty and prediction interval, \mbox{X-TRACK-MCD} yields a higher mean variance of $1.32$ and a wider prediction interval of $2.61$, compared with $0.75$ and $2.11$ for \mbox{X-TRACK-DE}, respectively. This shows that \mbox{X-TRACK-DE} gives narrower uncertainty regions even though the targeted and observed coverage are the same as \mbox{X-TRACK-MCD}. In terms of NLL, \mbox{X-TRACK-DE} achieves higher NLL $(0.51)$ as compared to \mbox{X-TRACK-MCD} $(0.47)$. After conformal prediction, both methods achieve approximately $95.5\%$ empirical conformal coverage, which closely aligns with the nominal target of $95\%$ at the evaluated horizon-wise marginal coverage level.
\begin{figure}[t]
    \centering
    \includegraphics[width=\columnwidth]{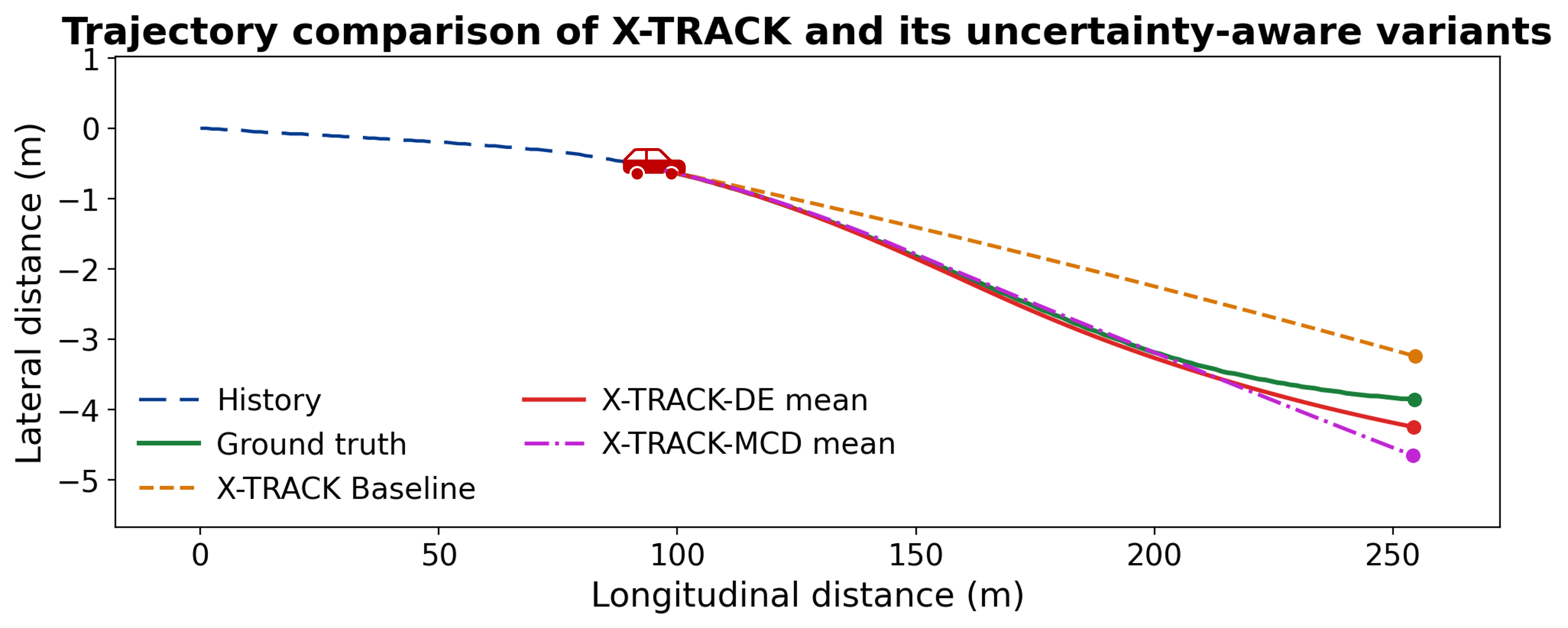}
    \caption{Qualitative comparison of ground truth with deterministic \mbox{X-TRACK} and predictive means of \mbox{X-TRACK-MCD} and \mbox{X-TRACK-DE}.}
    \label{fig:trajectory_comparison}
\end{figure}

\textbf{Qualitative Analysis.} Figure \ref{fig:trajectory_comparison} compares the ground truth trajectory with the trajectories generated by \mbox{X-TRACK} (baseline) and its uncertainty variants. Given a scenario, \mbox{X-TRACK-DE} (shown in red) yields the best trajectory. Here, \mbox{X-TRACK-MCD} outputs a predictive mean trajectory that closely aligns with the ground truth as compared to the baseline prediction.
\begin{figure}[t]
    \centering
        \includegraphics[
        width=\columnwidth
    ]{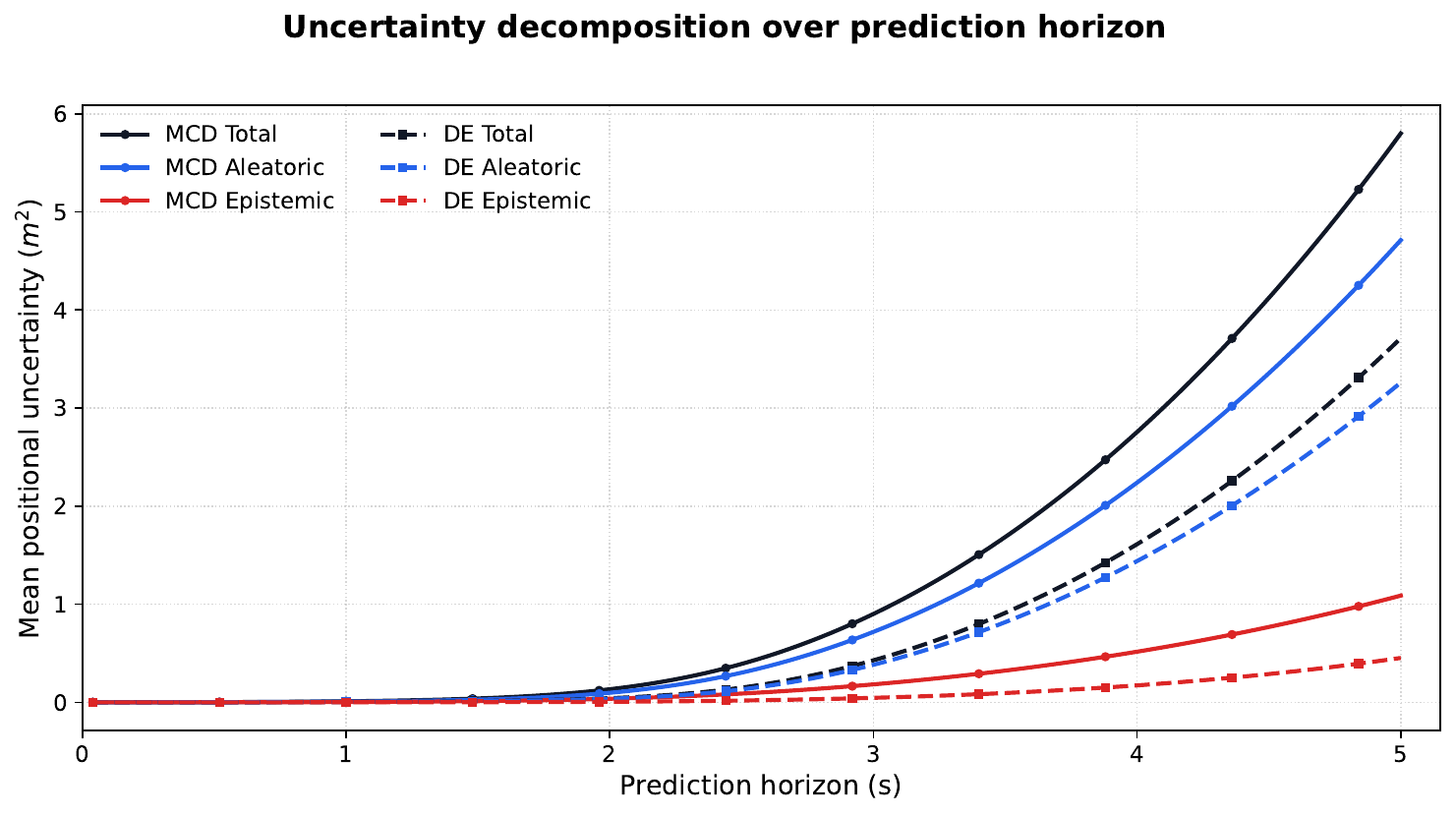}
    \caption{Decomposition of uncertainty into individual components for both variants.}
    \label{fig:uq_decomposition}
\end{figure}

Fig.~\ref{fig:uq_decomposition} shows the decomposition of the trajectory space uncertainty into aleatoric, epistemic, and total for X-TRACK-MCD and X-TRACK-DE over the predictive horizon. The aleatoric uncertainty has a significant contribution to the total, which increases nonlinearly with time for both methods. X-TRACK-MCD exhibits a stronger growth in both uncertainties, resulting in a larger total positional uncertainty at longer prediction horizons compared with \mbox{X-TRACK-DE}. 
\begin{figure}[t]
    \centering
        \includegraphics[
        width=0.80\columnwidth
    ]{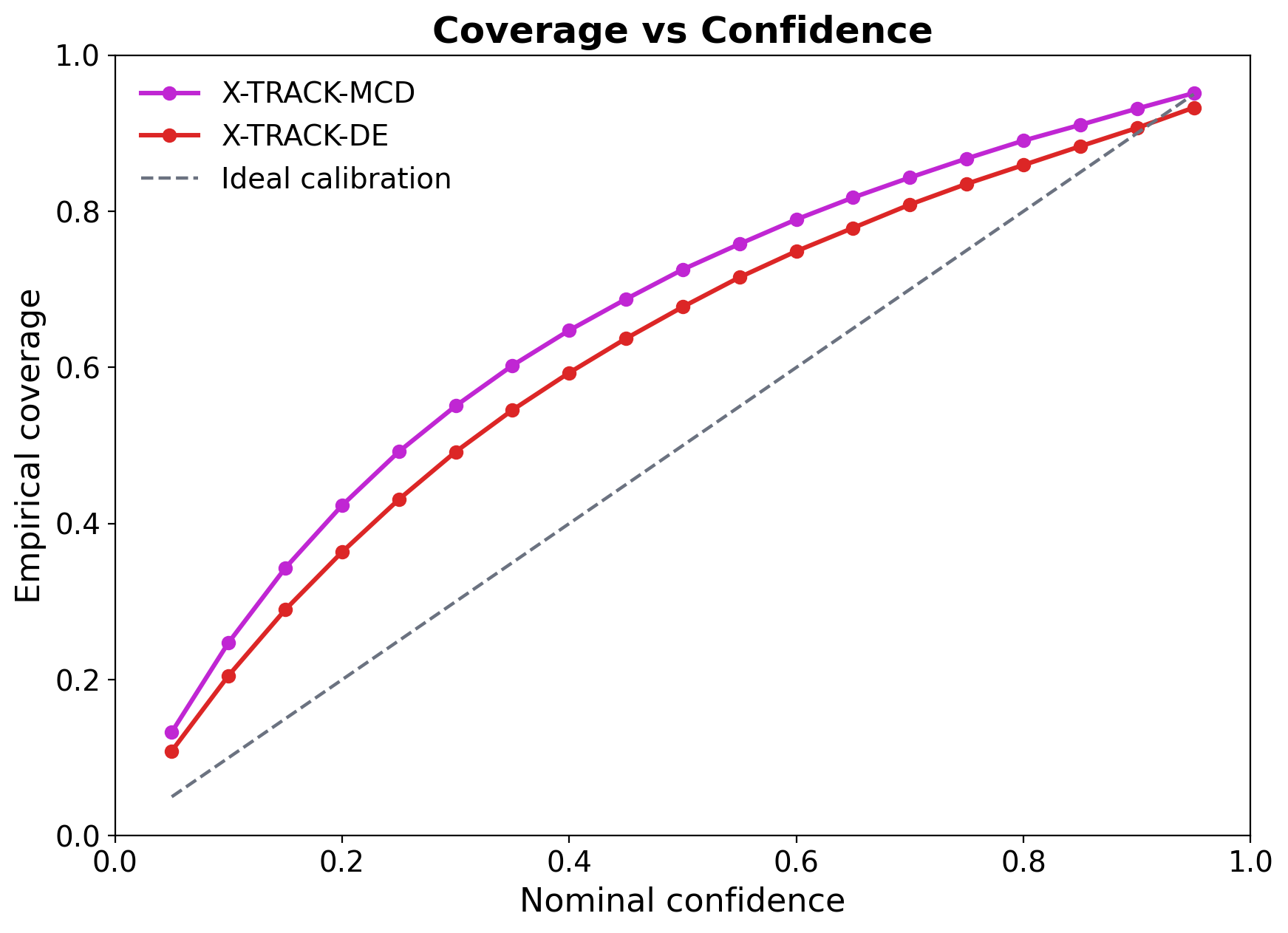}
    \caption{Comparing empirical coverage and nominal confidence for raw predictive distributions of \mbox{X-TRACK-MCD} and \mbox{X-TRACK-DE}.}
    \label{fig:coverage_plot}
\end{figure}

Fig.~\ref{fig:coverage_plot} shows the relationship between empirical coverage and nominal confidence level for both uncertainty-aware approaches. Across most confidence levels, both methods achieve empirical coverage above the ideal calibration line. As compared to X-TRACK-MCD, \mbox{X-TRACK-DE} remains closer to the calibration line over most of the confidence range. Both methods approach the ideal calibration behavior, as the nominal confidence approaches the target coverage level. Additional qualitative trajectory predictions, uncertainty regions, uncertainty with time, and sampled predictive trajectories are provided in the Appendix~\ref{sec:trajectory_comparison}.
\section{CONCLUSION}
This paper presented an uncertainty-aware extension of \mbox{X-TRACK} that incorporates predictive uncertainty into a physics-aware vehicle trajectory prediction framework. The proposed approach models uncertainty in the predicted vehicle motion variables and propagates it through the nonlinear kinematic rollout, rather than directly predicting or estimating uncertainty after obtaining the future position coordinates. Epistemic uncertainty is estimated using MC Dropout and Deep Ensembles, while aleatoric uncertainty is estimated using heteroscedastic Gaussian distributions over vehicle motion variables, longitudinal acceleration, and yaw rate. The estimated uncertainty is propagated from motion to trajectory space. Additionally, conformal prediction is applied to construct prediction regions with empirical coverage close to the nominal target.

Overall, within the considered highway trajectory prediction setting, \mbox{X-TRACK-DE} improves prediction accuracy while  \mbox{X-TRACK-MCD} mainly benefits shorter horizons. Both variants achieve empirical coverage close to the target level. Future work will focus on more diverse highway and traffic scenarios and improving conformal calibration methods.
\subsubsection*{ACKNOWLEDGMENT}
This research was supported by Technische Hochschule Augsburg and the Hightech Agenda Bavaria, funded by the Free State of Bavaria, Germany. The authors thank their colleagues at the Data Science und Autonome Systeme Technologietransferzentrum (TTZ) Landsberg for their insightful discussions and support.
\bibliography{references}
\clearpage
\appendix
\thispagestyle{empty}

\onecolumn
\aistatstitle{Supplementary Material}
\section{EXPERIMENTAL SETUP}
This section describes the evaluated baseline models, along with the training and implementation details provided to ensure reproducibility.
\subsection{Evaluated Baseline Models}
\label{sec:baseline}
Below are the models included in the comparison to evaluate the performance of the uncertainty-aware extension of \mbox{X-TRACK} with the competitive highway trajectory prediction baseline models.
\begin{itemize}
    \item \textbf{Multi-Head Attention LSTM (MHA-LSTM)} \citep{messaoud_attention_2021}: The model captures social interactions by using both global and local attention with an LSTM-based encoder-decoder model.
    \item \textbf{Intention-aware Non-Autoregressive Transformer (iNATran)}~\citep{chen_vehicle_2022}: To capture social and temporal dependencies, this model integrates GAT with a transformer (encoder) and temporal attention learning (TAL). The decoder combines cross-attention learning with intention-aware query generation.
    \item \textbf{Graph Fourier Transformation Neural Network (GFTNNv2)}~\citep{neumeier_prediction_2023}: Graph Fourier Transform (GFT) is used to transform vehicle interactions into a spectral scenario representation. By applying the inverse GFT, the prediction of the neural network is converted to the spatio-temporal domain.
    \item \textbf{Conditioned Vehicle Motion Diffusion (cVMD)}~\citep{neumeier_reliable_2024}: The approach employs diffusion models and a kinematic layer, along with uncertainty quantification, to improve prediction performance.
    \item \textbf{Uncertainty-Aware Conditioned Vehicle Motion Diffusion (cVMDx)}~\citep{neumeier_uq_2026}: An uncertainty-aware diffusion-based framework that represents predictive uncertainty using a Gaussian Mixture Model and efficiently generates multi-modal trajectories using Denoising Diffusion Implicit Model (DDIM) sampling.
    \item \textbf{xLSTM-based vehicle trajectory prediction (X-TRAJ)}~\citep{chugh_x-track_2026}: A variant of \mbox{X-TRACK} that predicts future vehicle positions directly in trajectory space without the physics-based kinematic rollout used in X-TRACK.
\end{itemize}
\subsection{Training and Implementation Details}
\label{sec:train_dets}
The model is trained for a maximum of $150$ epochs using a batch size of $32$ with the Adam optimizer~\citep{kingma_adam_2017} and an initial learning rate of $10^{-3}$. A multi-step learning rate scheduler is employed, reducing the learning rate by a factor of $0.1$ after $30$ and $60$ epochs. LeakyReLU with a negative slope of $0.1$ is used. Patience is set to $15$ to enable early stopping and terminate training when no improvement is observed for $15$ consecutive epochs. All the models are implemented in PyTorch $2.3$ using CUDA $12.4$ and trained on an NVIDIA L40S-4Q GPU with $4$ GB of GPU memory. Each model is trained independently on a single GPU.

The framework is trained using a Gaussian negative log-likelihood in both the trajectory and motion spaces. The trajectory loss $\mathcal{L}_{\mathrm{traj}}$ supervises the propagated trajectory distribution obtained through aleatoric sampling and kinematic rollout, while the dynamic loss $\mathcal{L}_{\mathrm{dyn}}$ supervises the predicted Gaussian distributions of vehicle motion variables, i.e., longitudinal acceleration and yaw rate. For each training sample, the overall loss is given by:
\begin{equation}
\mathcal{L}
=
\mathcal{L}_{\mathrm{traj}}
+
\alpha_\mathrm{dyn}
\mathcal{L}_{\mathrm{dyn}},
\end{equation}
where $\alpha_\mathrm{dyn}$ influences the contribution to the total NLL $\mathcal{L}$ coming from vehicle motion variables. In all experiments, $\alpha_\mathrm{dyn}=0.1$ is set through empirical evaluation balancing the influence of vehicle motion variables and trajectory accuracy. The contributions $\mathcal{L}_\mathrm{traj}$ and $\mathcal{L}_\mathrm{dyn}$ are defined on a per-sample basis as:
\begin{equation}
\mathcal{L}_{\mathrm{traj}}
=
\frac{1}{2t_f}
\sum_{t=t_{\mathrm{obs}}+1}^{t_{\mathrm{obs}}+t_f}
\sum_{d\in\{x,y\}}
\left[
\log\left(\sigma_{d,t}^{2}\right)
+
\frac{
\left(
p_{d,t}^{\mathrm{gt}}-\mu_{d,t}
\right)^2
}{
\sigma_{d,t}^{2}
}
\right],
\label{eq:traj_nll}
\end{equation}
\begin{equation}
\mathcal{L}_{\mathrm{dyn}}
=
\frac{1}{2t_f}
\sum_{t=t_{\mathrm{obs}}+1}^{t_{\mathrm{obs}}+t_f}
\sum_{d\in\{a_x,\dot{\psi}\}}
\left[
\log\left(\sigma_{d,t}^{2}\right)
+
\frac{
\left(
u_{d,t}^{\mathrm{gt}}
-
\mu_{d,t}
\right)^2
}{
\sigma_{d,t}^{2}
}
\right].
\label{eq:dyn_nll}
\end{equation}
During training, the per-sample losses are averaged over a mini-batch of size $B=32$. The resulting training objective is
\begin{equation}
\mathcal{L}_{\mathrm{batch}}
=
\frac{1}{B}
\sum_{b=1}^{B}
\left(
\mathcal{L}_{\mathrm{traj}}^{(b)}
+
\alpha_{\mathrm{dyn}}
\mathcal{L}_{\mathrm{dyn}}^{(b)}
\right),
\qquad B=32.
\end{equation}
\textbf{MC Dropout Settings.}
For MC Dropout, dropout is enabled during both training and inference. At multiple stages of the \mbox{X-TRACK} architecture, dropout is introduced, including the input embedding layer, xLSTM encoder, GAT layer, and decoder. The corresponding dropout probabilities are set to $0.15$, $0.20$, $0.20$, and $0.20$, respectively. Both the dropouts that are after the xLSTM encoder and the LSTM decoder are locked dropout~\citep{merity_regularizing_2018}, which means that a single dropout mask is sampled once and reused for the entire sequence rather than generating a new random dropout mask at every single time step. For each input, $M=20$ stochastic forward passes are performed, keeping dropout active during inference.

\textbf{Deep Ensemble Settings.}
Five independently trained \mbox{X-TRACK} models are used in this case, i.e., $K=5$. To introduce model diversity, random seed initialization and slight changes in network architecture are employed, as shown in Table \ref{tab:ensemble}, keeping the training process similar. The design choices associated with the deep ensembles are further described in additional ablation studies~\ref{sec:ensemble_ablation}.

\begin{table}[t]
\centering
\caption{Architectural modification of \mbox{X-TRACK} for the Deep Ensemble members.}
\label{tab:ensemble}
\begin{tabular}{lp{4.2cm}}
\toprule
\textbf{Model} & \textbf{Architectural Changes} \\
\midrule
Baseline & Default architecture \\
Wider Encoder & Encoder: $96$, Decoder: $160$, Dynamic Embedding: $96$ \\
Wider Embedding & Input Embedding: $48$, Dynamic Embedding: $80$ \\
More Heads & GAT Heads: $6$ \\
Deeper Decoder & Number of Decoder LSTM Layers: $3$, Dropout: $0.25$ \\
\bottomrule
\end{tabular}
\end{table}
\textbf{Aleatoric Uncertainty Settings.}
As the model predicts the mean and variance of longitudinal acceleration and yaw rate at each time step, $N=16$ vehicle motion variables are sampled from the predicted Gaussian distributions. These are propagated through the kinematic layer rollout to obtain future trajectories. For aleatoric and MC dropout, the total number of trajectories is $M \times N$, and for aleatoric and ensembles, it is $K \times N$.

\textbf{Conformal Prediction Settings.} The held-out calibration set contains $S_\mathrm{cal} = 685$ scenarios. The target coverage level is set to $95\%$, making $\delta = 0.05$. The conformal quantile is selected using the finite-sample rank $\lceil(S_{\mathrm{cal}}+1)0.95\rceil$. The dataset split is performed at the scenario level, and the held-out calibration set is kept strictly separate from model training and hyperparameter selection. The conformal calibration assumes exchangeability between the calibration and test scenarios. As a separate conformal quantile is estimated for each prediction time step, the resulting prediction regions provide horizon-wise marginal coverage under the exchangeability assumption between calibration and test samples. At each time step, a single conformal quantile is estimated using Mahalanobis scores computed from the $2\times2$ total positional covariance. 

\subsection{Code, Reproducibility and Asset Information}
\label{sec:reproduce}
The source code of the proposed uncertainty-aware \mbox{X-TRACK} framework will be made publicly available upon acceptance. The released repository will include the code for training and evaluating both \mbox{X-TRACK-MCD} and \mbox{X-TRACK-DE}. The README file will provide instructions for environment setup and dependencies, dataset preparation, model training, and evaluation, along with the configuration files required to reproduce the experimental setting reported in the paper. Appendix~\ref{sec:train_dets} provides details on the hyperparameters and training procedure.

\paragraph{Existing Assets.} Experiments are performed on the highD dataset~\citep{krajewski_highd_2018}, which, subject to the terms and conditions specified by its providers, is available free of charge for academic and research purposes. The dataset permits eligible non-commercial research use and restricts redistribution. Therefore, the highD dataset will not be included in the released code, and instructions for obtaining the dataset from the official provider will be included in the README file. 

The framework proposed in this paper extends X-TRACK~\citep{chugh_x-track_2026}. \mbox{X-TRACK} is distributed under the MIT License, and its implementation is used in accordance with this license. The uncertainty-quantification extensions introduced in this work will also be released under the MIT License.

\section{EVALUATION METRICS}
In this section, the widely used trajectory prediction and uncertainty quantification metrics are covered.
\subsection{Trajectory Prediction Metrics}
\label{sec:traj_pred_metrics}
\textbf{Average Displacement Error (ADE):} The average Euclidean distance across all time steps and all trajectories between the ground truth and predictive mean trajectories.
\begin{align}
\mathrm{ADE}
=
\frac{1}{S\,t_f}
\sum_{s=1}^{S}
\sum_{t=t_{\mathrm{obs}}+1}^{t_{\mathrm{obs}}+t_f}
\left\|
\hat{\mathbf{p}}_{t,s}
-
\mathbf{p}_{t,s}
\right\|_{2},
\label{eq:ade}
\end{align}
where $S$ denotes the number of test scenarios, $\hat{\mathbf{p}}_{t,s}$ is the predicted mean position for the test trajectory $s$ at time step $t$, and $\mathbf{p}_{t,s}$ is the corresponding ground truth position.

\textbf{Final Displacement Error (FDE):} The Euclidean distance, averaged across all trajectories, between the ground truth and predicted final positions for each predictive mean trajectory. 
\begin{align}
\mathrm{FDE}
=
\frac{1}{S}
\sum_{s=1}^{S}
\left\|
\hat{\mathbf{p}}_{t_{\mathrm{obs}}+t_f,s}
-
\mathbf{p}_{t_{\mathrm{obs}}+t_f,s}
\right\|_{2}.
\label{eq:fde}
\end{align}

\textbf{Root Mean Square Error (RMSE) at time t:} For every $S$ trajectories, the square root of the average of the squared differences between the predicted and the ground truth positions.
\begin{align}
\mathrm{RMSE}(t)
=
\sqrt{
\frac{1}{S}
\sum_{s=1}^{S}
\left\|
\hat{\mathbf{p}}_{t,s}
-
\mathbf{p}_{t,s}
\right\|_{2}^{2}
}.
\label{eq:rmse}
\end{align}
\textbf{Minimum ADE (minADE):} Among all predicted trajectories, minADE measures the smallest average displacement error.
\begin{align}
\mathrm{minADE}
=
\frac{1}{S}
\sum_{s=1}^{S}
\min_{q=1,\ldots,Q}
\left[
\frac{1}{t_f}
\sum_{t=t_{\mathrm{obs}}+1}^{t_{\mathrm{obs}}+t_f}
\left\|
\hat{\mathbf{p}}_{t,s}^{(q)}
-
\mathbf{p}_{t,s}
\right\|_{2}
\right],
\qquad Q=80.
\label{eq:minade}
\end{align}
where $Q$ is the number of sampled trajectories for each test scenario.

\textbf{Minimum FDE (minFDE):} The smallest final displacement error (FDE) across all the predicted trajectories.
\begin{align}
\mathrm{minFDE}
=
\frac{1}{S}
\sum_{s=1}^{S}
\min_{q=1,\ldots,Q}
\left\|
\hat{\mathbf{p}}_{t_{\mathrm{obs}}+t_f,s}^{(q)}
-
\mathbf{p}_{t_{\mathrm{obs}}+t_f,s}
\right\|_2,
\qquad Q=80.
\label{eq:minfde}
\end{align}
\textbf{Miss Rate (MR):} The percentage of predicted trajectories with final displacement errors (FDE) greater than a given threshold $d = 2\mathrm{m}$.
\begin{equation}
\mathrm{MR}
=
\frac{1}{S}
\sum_{s=1}^{S}
\mathbb{I}
\left(
\mathrm{minFDE}_{s} > d
\right),
\label{eq:mr}
\end{equation}
To ensure a fair comparison of both the uncertainty-aware variants, the number of sampled trajectories is set to $80$ for minADE, minFDE, and MR. As the total number of trajectories generated by MC dropout is $320$, $80$ trajectories are sampled from the total trajectories. 

Note that the predictive mean trajectory (defined in Eq.~\ref{eq:mean_traj}) is used for trajectory-error metrics, likelihood evaluation, conformal residual computation, and qualitative trajectory visualization. Due to the nonlinear kinematic rollout, this predictive mean generally differs from the deterministic mean-control trajectory defined in Eq.~\ref{eq:det_traj}.
\subsection{Uncertainty Evaluation Metrics}
\label{sec:unc_eval_metrics}
\textbf{Negative Log Likelihood (NLL):} Evaluates the likelihood assigned to the ground-truth trajectory under the predicted two-dimensional Gaussian distribution. Let
$
\mathbf{e}_{t,s}
=
\mathbf{p}_{t,s}
-
\boldsymbol{\mu}_{t,s},
$
the per-time-step NLL is
\begin{equation}
\ell^{\mathrm{NLL}}_{t,s}
=
\frac{1}{2}
\left[
2\log(2\pi)
+
\log
\det
\left(
\boldsymbol{\Sigma}_{\mathrm{total},t,s}
+
\epsilon\mathbf{I}
\right)
+
\mathbf{e}_{t,s}^{\top}
\left(
\boldsymbol{\Sigma}_{\mathrm{total},t,s}
+
\epsilon\mathbf{I}
\right)^{-1}
\mathbf{e}_{t,s}
\right],
\label{eq:nll}
\end{equation}

\begin{equation}
\mathrm{NLL}
=
\frac{1}{S t_f}
\sum_{s=1}^{S}
\sum_{t=t_{\mathrm{obs}}+1}^{t_{\mathrm{obs}}+t_f}
\ell^{\mathrm{NLL}}_{t,s}.
\label{eq:mean_nll}
\end{equation}
NLL is evaluated using the raw predictive covariance before conformal prediction. Lower values indicate that the predicted probabilistic distribution assigns higher likelihood to the observed trajectory.

\textbf{Confidence-level Coverage (C):} For a confidence level $c$, the squared Mahalanobis distance is
\begin{equation}
D^{2}_{t,s}
=
\mathbf{e}_{t,s}^{\top}
\left(
\boldsymbol{\Sigma}_{\mathrm{total},t,s}
+
\epsilon\mathbf{I}
\right)^{-1}
\mathbf{e}_{t,s}.
\label{eq:mahalanobis}
\end{equation}

For a two-dimensional Gaussian distribution, the corresponding
$\chi^{2}$ threshold with two degrees of freedom is
\begin{equation}
\tau_c
=
F^{-1}_{\chi^{2}_{2}}(c)
=
-2\log(1-c),
\qquad
c\in\{0.05,0.10,\ldots,0.95\}.
\label{eq:coverage_threshold}
\end{equation}

The empirical coverage is then
\begin{equation}
C(c)
=
\frac{1}{S t_f}
\sum_{s=1}^{S}
\sum_{t=t_{\mathrm{obs}}+1}^{t_{\mathrm{obs}}+t_f}
\mathbb{I}
\left(
D^{2}_{t,s}
\le
\tau_c
\right).
\label{eq:coverage}
\end{equation}

\textbf{Expected Calibration Error (ECE):}
Measures the discrepancy between nominal confidence levels and empirical coverage of the raw predictive distribution:
\begin{equation}
\mathrm{ECE}
=
\frac{1}{L}
\sum_{\ell=1}^{L}
\left|
C(c_{\ell})-c_{\ell}
\right|,
\qquad L=19.
\end{equation}
The coverage $C(c_{\ell})$ is computed from the full total positional covariance
using Eq.~\ref{eq:mahalanobis}. ECE is evaluated before conformal calibration.

\textbf{Pearson Correlation Coefficient ($\rho$):} Estimates the linear relationship between the predicted uncertainty and the average displacement error (ADE).
\begin{equation}
\rho_{\mathrm{ADE},u}
=
\frac{
\operatorname{Cov}(\mathrm{ADE},u)
}{
\sigma_{\mathrm{ADE}}\sigma_u
},
\label{eq:pearson}
\end{equation}
\begin{equation}
u_s
=
\frac{1}{t_f}
\sum_{t=t_{\mathrm{obs}}+1}^{t_{\mathrm{obs}}+t_f}
\operatorname{tr}
\left(
\boldsymbol{\Sigma}_{\mathrm{total},t,s}
\right),
\label{eq:sample_uncertainty}
\end{equation}
\begin{equation}
\mathrm{ADE}_s
=
\frac{1}{t_f}
\sum_{t=t_{\mathrm{obs}}+1}^{t_{\mathrm{obs}}+t_f}
\left\|
\hat{\mathbf{p}}_{t,s}
-
\mathbf{p}_{t,s}
\right\|_2,
\end{equation}
where $s = 1, 2, \ldots, S$ are the indexes of test scenarios. $\operatorname{Cov}(\mathrm{ADE},u)$ denotes the covariance between the trajectory error $\mathrm{ADE}$ and the predicted uncertainty $u$, and $\sigma_{\mathrm{ADE}}$ and $\sigma_u$ denote their standard deviations. The scalar uncertainty measure is therefore the trace of the raw total
predictive covariance, and is evaluated before conformal calibration.

\textbf{Mean Predictive Uncertainty:}
The average total positional uncertainty over all test scenarios and prediction
horizons is defined as
\begin{equation}
\bar{\sigma}^{2}
=
\frac{1}{S t_f}
\sum_{s=1}^{S}
\sum_{t=t_{\mathrm{obs}}+1}^{t_{\mathrm{obs}}+t_f}
\operatorname{tr}
\left(
\boldsymbol{\Sigma}_{\mathrm{total},t,s}
\right).
\label{eq:mean_uncertainty}
\end{equation}

\textbf{Conformal Prediction Coverage (CP Coverage):}
The empirical conformal coverage is evaluated using the conformalized total
covariance $\hat{\boldsymbol{\Sigma}}_{\mathrm{cp},t,s}$:
\begin{equation}
C_{\mathrm{cp}}
=
\frac{1}{S t_f}
\sum_{s=1}^{S}
\sum_{t=t_{\mathrm{obs}}+1}^{t_{\mathrm{obs}}+t_f}
\mathbb{I}
\left[
\mathbf{e}_{t,s}^{\top}
\hat{\boldsymbol{\Sigma}}_{\mathrm{cp},t,s}^{-1}
\mathbf{e}_{t,s}
\leq 1
\right].
\label{eq:cp_coverage}
\end{equation}

The coverage reported in Table~\ref{tab:uq_results} corresponds to empirical conformal coverage at the target $95\%$ level.

\textbf{Mean Prediction Interval Width (MPIW):}
Measures the average axis-aligned coordinate width of the conformal
prediction ellipse.

\begin{equation}
\mathrm{MPIW}
=
\frac{1}{2 S t_f}
\sum_{s=1}^{S}
\sum_{t=t_{\mathrm{obs}}+1}^{t_{\mathrm{obs}}+t_f}
\sum_{d\in\{x,y\}}
\left(
U_{d,t,s}
-
L_{d,t,s}
\right),
\label{eq:mpiw}
\end{equation}

where
\begin{align}
L_{d,t,s}
&=
\mu_{d,t,s}
-
\sqrt{
\left[
\hat{\boldsymbol{\Sigma}}_{\mathrm{cp},t,s}
\right]_{dd}
},
\\
U_{d,t,s}
&=
\mu_{d,t,s}
+
\sqrt{
\left[
\hat{\boldsymbol{\Sigma}}_{\mathrm{cp},t,s}
\right]_{dd}
}.
\end{align}

\section{ABLATION STUDIES}
This section presents additional ablation studies conducted to evaluate the impact of key design choices of the proposed framework.

\subsection{Homogeneous and Heterogeneous Deep Ensembles}

\label{sec:ensemble_ablation}
To understand the influence of ensemble diversity on predictive and epistemic uncertainty, additional experiments are carried out. A homogeneous ensemble uses identical X-TRACK architectures trained with different random seeds, and a heterogeneous ensemble uses additional minor architectural variations, as described in Appendix~\ref{sec:train_dets}. Both variants estimate only the epistemic uncertainty, without aleatoric modeling or conformal calibration.

The heterogeneous ensemble improves both trajectory prediction and epistemic uncertainty estimation as compared to the homogeneous ensemble. In the case of heterogeneous ensembles, ADE improves from $0.65$ to $0.58$ and FDE from $1.85$ to $1.72$. In addition to this, the empirical pre-CP coverage increases from $25.2\%$ to $42.4\%$, while ECE decreases from $0.25$ to $0.08$. However, since the heterogeneous ensemble members also differ in model capacity, this ablation does not isolate architectural diversity as the sole source of these improvements.

However, the empirical coverage is substantially below the target $95\%$ coverage level, indicating that both ensembles remain overconfident. This motivates the modeling of aleatoric uncertainty and conformal calibration. 
\begin{table}[t]
    \centering
    \caption{Comparison of homogeneous and heterogeneous ensembles to assess the impact of architectural diversity.}
    \label{tab:ensemble_ablation}
    \begin{tabular}{lcccc}
        \toprule
        Architecture & ADE $\downarrow$ & FDE $\downarrow$
        & Pre-CP C@95\% $\uparrow$ & ECE $\downarrow$ \\
        \midrule
        Homogeneous Ensemble
        & 0.65 & 1.85 & 25.2\% & 0.25 \\
        Heterogeneous Ensemble
        & \textbf{0.58} & \textbf{1.72}
        & \textbf{42.4\%} & \textbf{0.08} \\
        \bottomrule
    \end{tabular}

    \vspace{2pt}
    \footnotesize
    \textit{Note:} Both variants estimate only epistemic uncertainty. The reported coverage is evaluated before conformal prediction.
\end{table}

\subsection{Analyzing the Effect of Conformal Prediction}
To investigate the impact of applying conformal prediction to the underlying uncertainty estimation framework, Table~\ref{tab:cp_ablation} shows the comparison of empirical coverage of the raw total predictive covariance with the coverage obtained after conformal prediction. Using the held-out calibration set, conformal prediction is applied to the total $2\times2$ positional covariance.

Conformal prediction substantially reduces the deviation between empirical and nominal coverage for both approaches (X-TRACK-DE and X-TRACK-MCD). After calibration, it was observed that the empirical coverage closely matches the target $95\%$ level, indicating that this step calibrates the raw predictive uncertainty.
\begin{table}[t]
\centering
\caption{Effect of conformal prediction on the uncertainty-aware variants.}
\label{tab:cp_ablation}
\begin{tabular}{lcccc}
\toprule
& \multicolumn{2}{c}{X-TRACK-MCD}
& \multicolumn{2}{c}{X-TRACK-DE} \\
\cmidrule(lr){2-3}
\cmidrule(lr){4-5}
Metric & Pre-CP & Post-CP & Pre-CP & Post-CP \\
\midrule
Coverage@95\%  $\uparrow$ & 67.47\% & 95.72\% & 62.01\% & 95.39\% \\
MPIW $\downarrow$ & 2.24 & 2.61 & 1.57 & 2.11 \\
ECE $\downarrow$ & 0.17 & 0.11 & 0.12 & 0.11 \\
\bottomrule
\end{tabular}
\end{table}

\section{ADDITIONAL QUALITATIVE TRAJECTORY PREDICTION RESULTS}
\label{sec:trajectory_comparison}
\begin{figure}[H]
    \centering
        \includegraphics[
    width=\textwidth,        trim=0 0 0 0,
        clip]{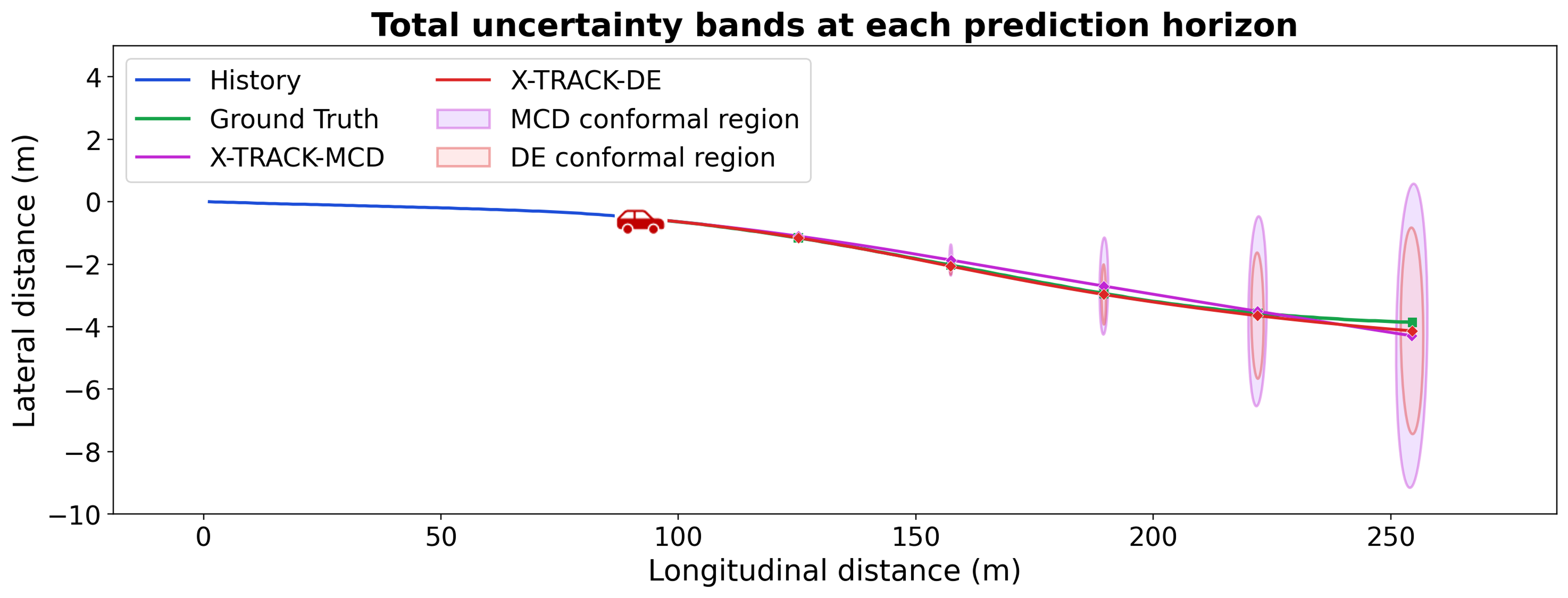}
    \caption{Comparison of total predictive uncertainty for \mbox{X-TRACK-MCD} and \mbox{X-TRACK-DE} over different prediction horizons.}
    \label{fig:uncertainty_band}
\end{figure}
Figure~\ref{fig:uncertainty_band} illustrates the total uncertainty in trajectory space for both \mbox{X-TRACK-MCD} and \mbox{X-TRACK-DE} at different prediction horizons. The plot confirms that the uncertainty regions are compact at shorter horizons, while the regions expand with the prediction horizon. Qualitatively, for the given scenario, \mbox{X-TRACK-MCD} exhibits wider uncertainty regions than its DE counterpart. Both the predictive mean trajectories remain close to the ground truth trajectory despite increasing uncertainty and seem to diverge only at the $5s$ prediction horizon.

\begin{figure}[H]
    \centering
    \includegraphics[width=\textwidth]{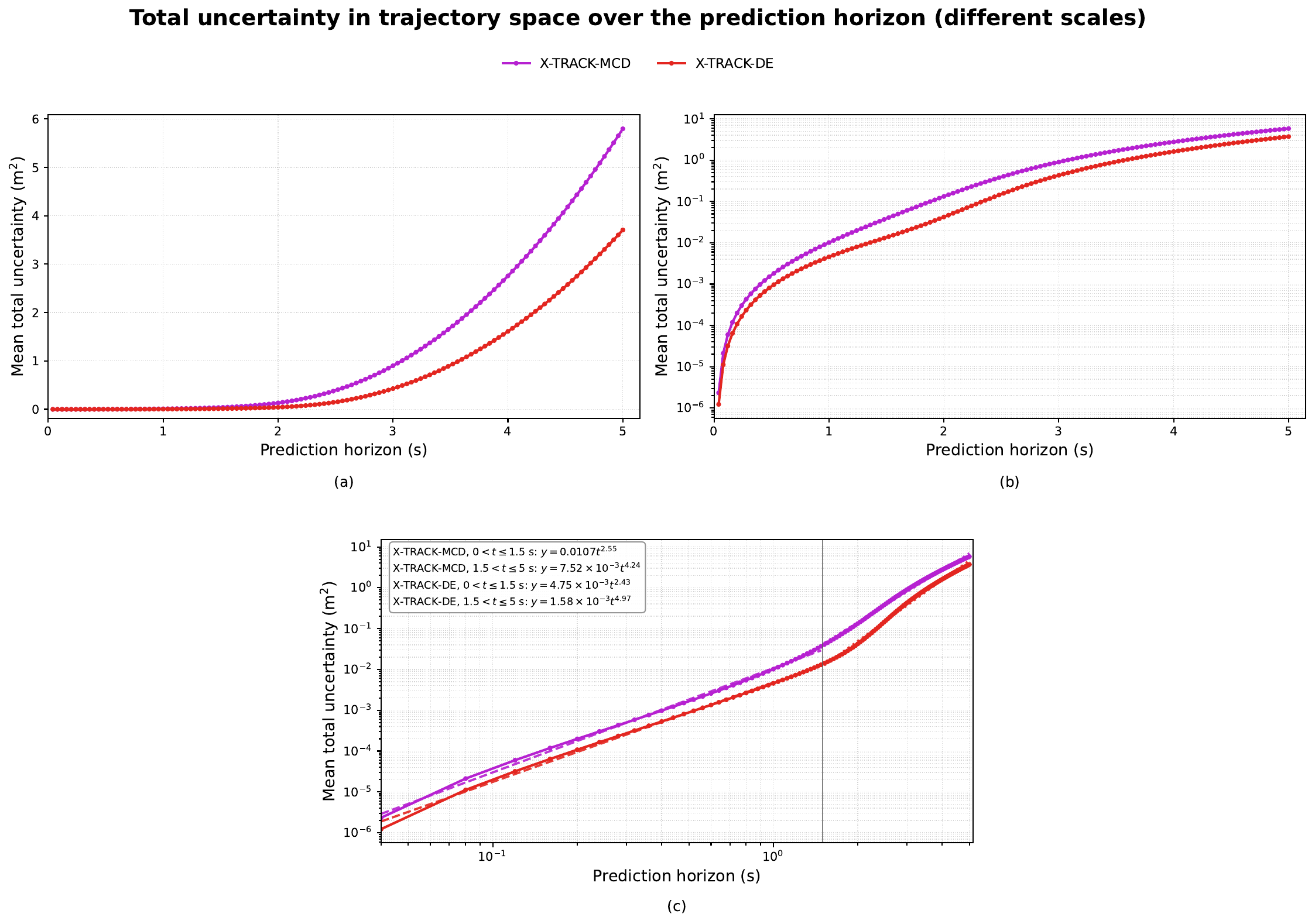}
    \caption{Total predictive uncertainty in trajectory space over the prediction horizon for X-TRACK-MCD and X-TRACK-DE. The same mean uncertainty is shown using (a) linear, (b) semi-logarithmic, and (c) log-log scales.}
    \label{fig:uq_log_scale}
\end{figure}
Figure~\ref{fig:uq_log_scale} presents the total trajectory space uncertainty with respect to the prediction horizon using linear, semi-logarithmic, and log-log scales. The logarithmic representations show that the predictive uncertainty spans several orders of magnitude and increases nonlinearly with the prediction horizon, with comparatively small uncertainty at shorter horizons followed by a substantially stronger increase at longer horizons. This nonlinear increase is consistent with the accumulation and propagation of motion space uncertainty through the iterative kinematic rollout.

\begin{figure}[H]
    \centering
        \includegraphics[
            width=\textwidth,
    ]{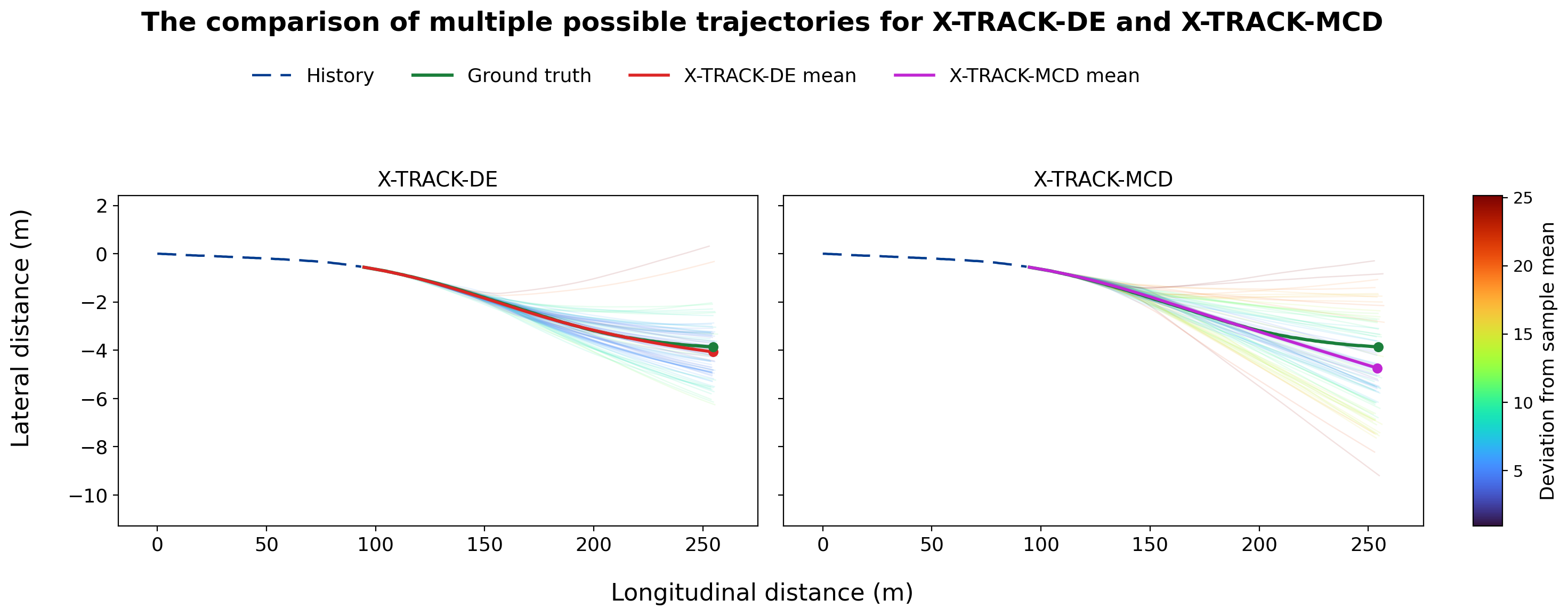}
    \caption{Comparison of predictive multiple trajectories obtained by \mbox{X-TRACK-DE} and X-TRACK-MCD.}
    \label{fig:sampled_trajectories}
\end{figure}

Figure~\ref{fig:sampled_trajectories} represents the predictive trajectory distributions obtained using \mbox{X-TRACK-DE} and \mbox{X-TRACK-MCD} for a particular traffic scenario. The multiple possible trajectories ($Q = 80$) correspond to different aleatoric samples, along with different ensemble predictions or stochastic MC dropout forward passes. The predictive mean of both approaches follows the ground truth closely and diverges progressively at longer horizons. The color of each trajectory represents its deviation from the corresponding predictive mean.

\begin{figure}[H]
    \centering
        \includegraphics[
        width=\textwidth,
    ]{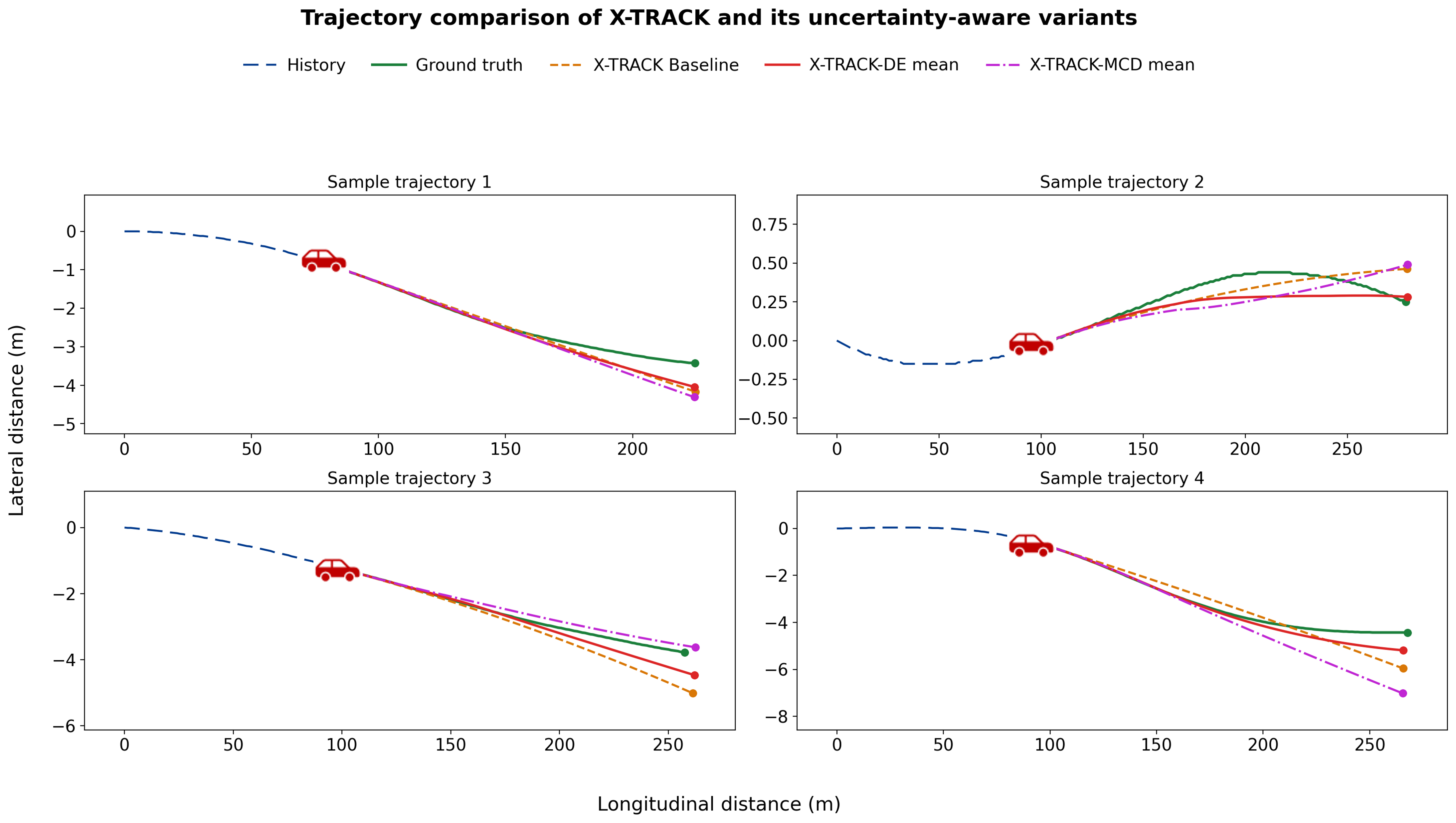}
    \caption{Qualitative comparison of ground-truth trajectories with deterministic \mbox{X-TRACK} and the predictive means of its uncertainty-aware variants.}
    \label{fig:trajectory_comparison_plot}
\end{figure}

Figure~\ref{fig:trajectory_comparison_plot} shows a qualitative comparison of the ground truth trajectory along with the prediction obtained using deterministic \mbox{X-TRACK} and its uncertainty-aware variants based on Deep Ensembles (X-TRACK-DE) and Monte Carlo Dropout (X-TRACK-MCD). The predicted trajectories seem to diverge toward the end of the prediction horizon, and in general all model variants follow the right direction of the future vehicle motion. The sample trajectories from the test set illustrate the different trajectories predicted by the deterministic as well as uncertainty-aware approaches despite having the same underlying \mbox{X-TRACK} backbone.

\begin{figure}[H]
    \centering
        \includegraphics[
        width=\textwidth,
    ]{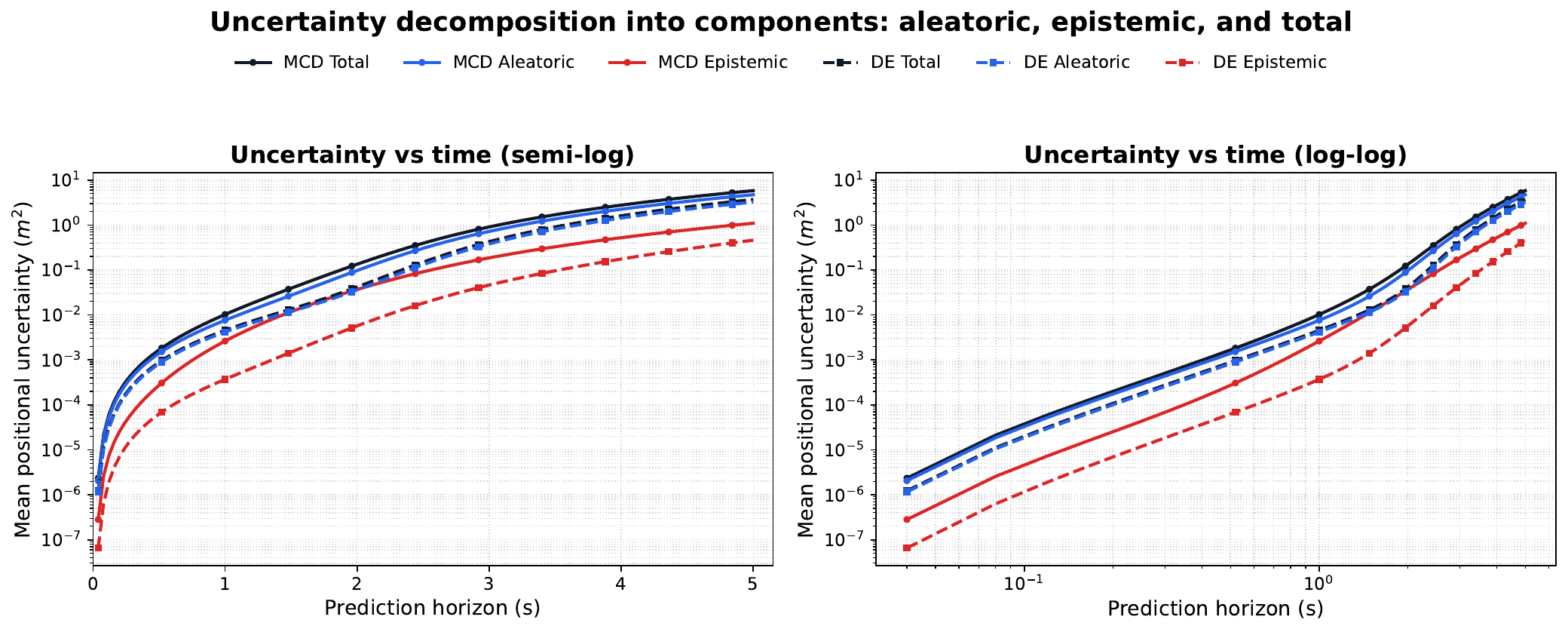}
    \caption{Decomposition of trajectory space predictive uncertainty (different scales) into aleatoric, epistemic, and total components for X-TRACK-MCD and X-TRACK-DE.}
    \label{fig:uq_decomposition_scales}
\end{figure}

Fig.~\ref{fig:uq_decomposition_scales} shows the uncertainty decomposition on semi-logarithmic and log-log scales as an extension of figure~\ref{fig:uq_decomposition}. The semi-logarithmic representation highlights the increasing rate of uncertainty growth at longer prediction horizons. In the log-log representation, the varying slope suggests that the growth of the different uncertainty components is not well described by a single power law over the entire prediction horizon, with stronger growth at longer horizons.
\end{document}